B.Sc. in Computer Science and Engineering Thesis

# A Temporal Psycholinguistics Approach to Identity Resolution of Social Media Users

Submitted by

Md. Touhidul Islam
201205031

Supervised by

Dr. Mohammed Eunus Ali

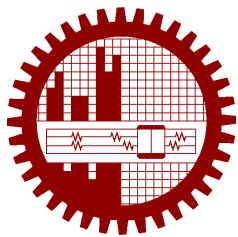

**Department of Computer Science and Engineering**
**Bangladesh University of Engineering and Technology**

Dhaka, Bangladesh

September 2017

# CANDIDATES' DECLARATION

This is to certify that the work presented in this thesis, titled, "A Temporal Psycholinguistics Approach to Identity Resolution of Social Media Users", is the outcome of the investigation and research carried out by us under the supervision of Dr. Mohammed Eunus Ali.

It is also declared that neither this thesis nor any part thereof has been submitted anywhere else for the award of any degree, diploma or other qualifications.

---

Md. Touhidul Islam
201205031

# CERTIFICATION

This thesis titled, **"A Temporal Psycholinguistics Approach to Identity Resolution of Social Media Users"**, submitted by the group as mentioned below has been accepted as satisfactory in partial fulfillment of the requirements for the degree B.Sc. in Computer Science and Engineering in September 2017.

**Group Members:**

    **Md. Touhidul Islam**

**Supervisor:**

---
Dr. Mohammed Eunus Ali
Professor
Department of Computer Science and Engineering
Bangladesh University of Engineering and Technology

# ACKNOWLEDGEMENT

First of all, we would like to thank our honorable supervisor Professor Dr. Mohammed Eunus Ali for his ongoing support in our thesis work and all related research, for his enthusiasm, tolerance, and vast knowledge. This thesis would not have been possible without his constant motivation and we could not have hoped for a better thesis supervisor. We would like to express our gratitude to Md. Hasan Al Maruf, Graduate Research Assistant, University of Michigan for his continuous help in the proceedings of the thesis work, for the data collection, and algorithm design stage. We want to thank all the lab assistants of the Samsung lab for their help and support. We also thank all of our classmates who have been very helpful in making this thesis a success. Finally, we would like to express our profound gratitude to our family members. Without them, nothing of this would have been possible.

Dhaka                                                                                                    Md. Touhidul Islam
September 2017

# Contents

# List of Figures

# List of Tables

# List of Algorithms

# ABSTRACT

With the increasing number of social media sites around the web, people are more likely to have accounts in multiple sites. As a result, matching their accounts across several sites has become a problem of interest. The task of analyzing and comparing more than one profiles to find match among them in order to resolve identities is known as identity resolution. In this thesis, we build an identity resolver model that extracts topics, emotions, and timings of user interaction in different social media. The contents of the posts that people make on their social media accounts are vastly dependent on the timings. For this reason, we call it temporal behavior; signifying the importance of time behind the contents of the post. In our thesis, we build models both temporal (time dependent) and non-temporal (time independent) and make comparisons between them. In temporal models, we divide the whole timespan into some smaller time bins and for non-temporal models, we consider the whole timespan one big bin. One of our temporal models consists of two phases where the first phase is topic frequency analysis and second phase is emotion analysis. For our data, we use the publicly available posts of disqus and twitter of different users. We also build a distance based scoring model and monitor the performance. In our analysis, the temporal model is roughly better than the non-temporal model.

# Chapter 1

# Introduction

The number of social media sites around the web is ever increasing. At the same time, the number of accounts different people have in these sites also increases. According to [2], about $2.46$ billion people all over the world uses different social media sites in 2017. Table 1.1 shows the number of users in different social media sites. Many users have accounts in more than one social media site. According to [3], the average number of accounts per person is 7. To predict the behavior of a person under various circumstances and to find the other accounts in different sites, we can use the information provided in one of the social media profiles. This is the main idea of identity resolution in social media. Identity resolution is the process of searching and analyzing between disparate data sets and databases to find a match and resolve identities. A virtual profile is the collection of attributes that we can extract analyzing a user's profile. Identity resolution and virtual profile generation is useful in many aspects.

| Social Media Site | Number of Users (in millions) |
|:---:|:---:|
| Facebook | 1,871 |
| Whatsapp | 1,000 |
| Facebook Messenger | 1,000 |
| QQ | 877 |
| WeChat | 846 |
| QZone | 632 |
| Instagram | 600 |
| Twitter | 317 |
| Snapchat | 300 |
| Skype | 300 |
| Viber | 249 |
| LinkedIn | 106 |
| Disqus | 35 |

Table 1.1: Scoial media and users (According to [1], in January, 2017 )

## 1.1 Real Life Usage Scenarios

i. **Account Synchronization:** Nowadays, people have accounts in almost all the social media sites. Now, if they provide information in one site say, Facebook, that information will not be available in his other social media accounts unless we make some kind of linkup among them. This linkup is done through identity resolution.

ii. **Predicting Future Behaviors:** A user's comprehensive virtual profile can be a good tool to predict his behavior under different times and circumstances. Most of the people tend to follow a regular lifestyle. For example, someone with record of previous cyber crime have higher tendency to do that again, a person known for writing productive and useful things also tend to continue that. Thus, future behavior prediction can be done through a virtual profile generation.

iii. **Customer Service:** The quality of customer service can be enhanced. Insurance companies can do better risk management with the analysis of their policy holder's virtual profile.

iv. **Use in Ad-agencies:** Ad-agencies can find out their targets more easily and put the ads in correct places [4]. The interest of people tends to vary a lot from person to person. Some people may have a lot of interest in new smartphones, some may be interested in cars, motorcycles while others may be interested in computer accessories. A virtual profile provides us the information of one's things of interest. Thus, when placing ads to a person's homepage, ideal recommendations can be made.

v. **Fake Account and Rumor Detection:** Identity resolution can be crucial in security purposes like fake account identification [5], background check and rumor detection [6–9]. One of the major issues in today's social media is the existence of fake accounts. When a person uses fake data to create an account in a social media site, we call it fake account. Sadly, there is no direct way to solve this problem. But, a person's behavior can be analyzed from his virtual profile. And, if these behaviors end up matching quite some extents to a suspected fake profile, we can be highly confident that this fake account belongs to that person.

Rumors are bad in any forms whether in real life or social media. And, in a time when people pass most of their time in social media, social media rumors can have devastating effects on the society and country. Analyzing virtual profiles can help us in detecting rumors and take necessary actions.

## 1.2 Motivation

Our principal motivation behind this thesis is to build an identity resolver that is more general, more user friendly and most importantly, more accurate. The underlying topic, emotion, vibe of a post made by a user will play a vital role very in our approach to the identity resolution problem. A person who loves and follows football is likely to make posts about recent matches taking place around the world. Analyzing one's posts, we can find out one's emotion, attitude towards a team or a player. A *Barcelona* fan is likely to make a joyous or angry post depending on whether the team won or lost. A *Roger Federer* fan will try to uphold the quality of the favorite player. If one hates *Game Of Thrones*, *The Shawshank Redemption* or *Harry Potter*; one may make posts criticizing them. We can expect this person to make same type of posts or comments in the other social media accounts. Thus, we are extracting a lot of information that are not available in the basic profile. Combining all of these with the time of the posts can make a very robust profile matching technique.

## 1.3 Problem Definition

After presenting the definitions of identity resolution and its usage in real life, now we can state the problem we are trying to solve. Our target is to build a model that can match profiles across web platforms. For example, if we are given a user's Twitter profile and a probable list of Disqus profiles, our task is to identify the Disqus profile using the data available in both Twitter and Disqus profiles. We have to collect the data from user interaction. By user interaction, we means the textual contents available in their accounts.

The activities of person or the posts made by a person in social media is heavily dependent on the timing of their posts. This is where temporal behavior modeling will come in. Temporal behavior is the behavior that points to the activities of a person dependent on time. Most of the people stick to a steady lifestyle which means, a person is more likely to do a particular task in a particular time rather than in some random time of the day. We can find out the timings of one's posts in one of the social media accounts and use that to find one's other accounts in different sites. Let us say, we have a person who posts a *good morning* message, stays inactive during almost the whole day, and becomes active with posts at nights. What we can deduce from these information are this person probably is an early riser, probably passes a busy day with work, education or anything and, a less busy night with enough time to make posts.

## 1.4 Solution Overview

We plan to build several models in trying to solve th identity resolution problem. The models and their short description are given below.

i. **Topic Based Non-temporal Model:** Just like any non-temporal model, we do not consider timings here. We simply perform analysis based on topic frequency. The whole timespan is considered just one big bin.

ii. **Topic Based Temporal Model:** Just like any temporal model, we consider the timings here. The whole timespan is divided into some smaller overlapping time bins. Within this smaller time bins, an analysis is performed. Then, all the results of all bins are integrated to find the final output.

iii. **Sentiment Based Temporal Model:** This model is based on only sentiment. We divide the whole timespan into smaller overlapping bins, perform Sentiment Analysis within these smaller windows and then integrate to get the final result.

iv. **Two Phase Method:** This method is a temporal one. In this method, there are two phases. The first stage is a Topic Based Temporal Method. After the first stage, first $N$ profiles are selected and passed though a second stage of Sentiment Based Temporal Model. The later stage generates the final output.

v. **Distance Based Scoring Model:** This method of scoring is quite different from the above mentioned methods. This is a reward and punishment kind of method. Here, we give bonus or penalize different topics based on their frequencies.

Detail analysis of these models is presented in Chapter 4.

## 1.5 Challenges

To design a good and efficient identity resolution model, we will face the following challenges:

i. **Finding Enough Data:** To test whether a model actually works or not we need a massive amount of data. To find such huge amount of data can be tricky.

ii. **Reducing Complexity:** Complex models means higher computational time which we cannot always afford. Thus, we have to reduce the model complexity as much as we can.

iii. **Using Existing Resources:** Many resources are already in use regarding identity resolver models. But, they may not be directly suitable for us to use. We have to change them to our suit to use them when needed.

**iv. Finding Reliable Data:** Social media are full of fake accounts, sarcasms and unrealistic textual contents. If these are not handled separately, we can run into trouble building our target model.

## 1.6 Contribution

We have made the following contributions in this thesis:

**i.** For the first time, we have made topic extraction from the textual contents available in social media accounts of different users.

**ii.** We have extracted sentiment lying underneath plain texts of a post.

**iii.** We have built temporal models in which the whole time-space is divided into several time bins.

**iv.** We have built non-temporal models and compared their performance with the temporal models.

**v.** We have used dataset from Twitter and Disqus to do our analysis.

## 1.7 Thesis Organization

In the rest of this book, we present the existing works and the difference between our work and those in Chapter 2. In Chapter 3, we present our way of collecting, storing and filtering the data we used in our analysis. Chapter 4 describes the models we built in our experiment. Chapter 5 describes another distance based scoring model that we built. Chapter 6 includes all the testing results and analysis of those results where Chapter 7 contains the concluding remarks and future directions.

# Chapter 2

# Background and Related Works

## 2.1 A Brief History

Identity Resolution in Social Media is a relatively new topic. But the same is not true for only Identity Resolution. Identity resolution also goes by few other names like data deduplication [10], name matching [11], record linkage [12]. Identity resolution was first introduced in a paper called "Record linkage" by Hilbert L. Dunn in 1946 [13]. In 1969, Fellegi and Alan Sunter invented probabilistic identity matching frameworks [14] that are even in use now. The first step of identity matching is the collection of profile attributes. Preparation of data comes first when we consider identity matching. Data can come from databases, files, social media websites or any other sources. For example, the attribute *name* can be labeled as simply Name in one data source, and can have multiple parts in another source like, *first name, middle name, and last name*. Importantly, data sources may have different format for the value of a profile attribute.

Facebook, Linkedin, and Twitter usually have a consistent structure of attributes that are extracted by proposed framework. There can be two approaches towards identity resolution. One is called domain knowledge approach and the other is called probabilistic machine learning approach. Domain knowledge approach is done by use of comparison rules. Both the approaches have their own benefits and limitations.

## 2.2 Existing Methods

Researchers have employed techniques based on profile attributes to search for a user's account on a different site when one site's account is given [15–18].

### 2.2.1 Syntactic Matching

Researchers have tried syntactic comparison [15, 18–20]. Many papers have tried a syntactic profile comparison while matching different accounts in different websites. Motoyama and Varghase [15] tried syntactic identity resolution method by using MySPace [21] profile to find similar facebook profiles using the facebook search API [22].

Martin Szomszor et al. [23] performed profile matching across Delicious [24] and Flickr [24], two social media sites. Tereza Iofciu et al [25] used social tags and user Identities to match profiles among Flickr, Delicious, and Stumpleuopn [26], and obtained an amazing accuracy of upto 90%. Danesh Irani et al. [18] used names and nicknames to construct social media profiles of users.

Cunet T Akcora and Barbara Carminati [27] proposed a new method of guessing values for missing user profile attribute based on most common value they found among the user's online friends. Perito et al [19] showed that *username* attribute alone can be used to link profiles from separate social media platforms.

Vosecky et al [28] introduced a vector based comparison method for identity resolution in social media. He constructed a vector using all the profile attributes available in a user profile. Each attribute here was an element of the vector. When matching against other profiles, each corresponding attribute was compared and given a similarity score of 0 to 1. All the attributes had a different weight. Individual similarities were multiplied by their corresponding weights. Later, their summation corresponded to the final match score.

Vosecky et al [28] also introduced three different methods to compare attributes namely exact matching, partial matching, and fuzzy matching. Exact matching was used for comparing gender. Partial matching was used for comparing parts of profile attributes in cases of an abbreviation (Ex. : MD) or multiple terms (Ex. : Pattern Manager) etc. Fuzy matching was used in another name called VMN which was used to provide an approximate similarity value of two strings.

Vosecky et al [28] compared only profile attributes and does not take into account user's posts and network elements.

### 2.2.2 Semantic Matching

Researchers also tried to find the semantic match [29, 30]. There are many papers which tried to use different semantic string matching techniques in order to compare and match profiles. Cortis et al. [31] experimented with DBPedia [32] in an attempt to perform semantic matching across social media platforms. DBpedia [32] provides various semantic services based on the information available in wikipedia [33].

Goldbeck and Rothstein [30] proposed a technique that performs semantic profile matching using publicly available friends of friends [34] information from various blogs and social media sites. E. Raad et al [29] performed a heuristic procedure on the data and compared them with the available profile information using weighted similarity score.

### 2.2.3 Other Methods

while few researchers tried to combine several methods to obtain better results [20, 35], others tried new methods like temporal behavior analysis [35]. Each of these ideas has its own advantages and limitations. For example, people are not bound to provide correct information in their social media accounts. So, a model built based on syntactic profile attribute analysis can be completely misleading in such a case. Again, the attributes that are available in social media site is not necessarily available in other sites. So, a comparison can be invalid here.

### 2.2.4 Combining Several Methods

Some researchers used only one idea (only syntactical, only semantic, only temporal) [15, 18–20, 29, 30] while trying to make the match where other researches incorporated two or more ideas to do the profile matching [20, 35]. In most cases, the technique that used more than one idea got better results and efficiency [20, 35]. This is our main motivation behind our idea of incorporating temporal behavior analysis with the underlying topic, emotion, and vibe.

Now is the time to have a closer look at the researches that worked on the same topic with a slightly different approach.

## 2.3 Combined Temporal and Linguistic Model

Soroush Vosoughi et al. [35] experimented with some linguistic models as well as some temporal models. The best temporal model they used produced an accuracy of $10\%$ where their best linguistic model produced a $27\%$ accuracy. A combination of these models produced a maximum of $31\%$ accuracy in finding the correct account using the given one. For analysis, they used the linguistic contents and time of the posts made by user in social media accounts. The linguistic models they used were mostly based on syntactic comparison. Although temporal models, by themselves seem to be under-performing, [35] showed that combining temporal models with linguistic ones can actually produce better results.

## 2.4 Identity Search and Identity Matching

Paridhi Jainy et al. [20] decomposed the whole process of identity resolution into two tasks. *Identity search* and *Identity Matching*. *Identity search* dealt with the portion where they searched for a probable account in a site where *Identity Matching* managed the process of finding the right profile from a given list. For *Identity search*, they used four different methods namely *Profile Search*, *Content Search*, *Self-mention Search* and, *Network Search*. Once the probable list was generated, the *Identity Matching* portion came into action. *Identity Matching* included two techniques : syntactic matching and image matching. The matching process gave a score to each of the candidates residing in the probable list. Then they produced an ordered list based on the score.

## 2.5 Summary

We are going to use the contents and times of the posts made by a user in a social media site. So, we can say that our data collection will be somewhat similar to Soroush Vosoughi et al. [35]. What they did in their linguistic model was a syntactic comparison. But, we are going to analyze the emotion, feelings, vibe, and, topic lying underneath simple texts in the posts of a user. So, the two lines : *I hate Manchester United* and *I hate Manchester City* were quite close in their models where they will be completely different in ours. Again, they used *Facebook* and *Twitter* as their sources, we will use *Twitter* and *disqus*. Now to summarize, both the models used syntactical comparison in their process of identity resolution. But, we will capture the underlying emotion of the posts.

# Chapter 3

# Datasets

## 3.1 Data Collection and Filtering

### 3.1.1 Collection

No two particular websites seemed obvious choices to us for analysis. So, we started by using the *Disqus API* [36] to crawl through around $250,000$ profiles. What we basically searched in this step were the links to other social media sites (*Facebook*, *Twitter*, *Youtube*, *Google Plus* etc.). We found $4,331$ profiles crawling through *disqus* linking directly to their corresponding *twitter* counterpart. Thus, we had $4,331$ *disqus* profiles and their *twitter* counterparts. Let us call each *disqus* profile and its corresponding *twitter* profile together a **set**. We decided to work with these $4,331$ sets.

### 3.1.2 Storing

In the second stage, we used both the *Disqus API* [36] and the *Twitter API* [37] to collect data from the above mentioned sets. In case of *disqus*, we collected each of the comments, time of those comments, ID and time of the parent post etc. In crawling the *twitter* profiles, we collected only the textual content of the post and its time. We saved our collected in data in *json* format files for ease of later use.

### 3.1.3 Filtering

Later, we ran those sets of profile through some filtering processes. Initially, if a set contained a *disqus* or *twitter* profile with no posts at all; it was discarded. In further filtering, we also discarded those sets which contained a *disqus* or *twitter* profile with less than 20 posts because too little amount of posts have a tendency to produce wrong results. In the end, we had $2,525$ sets of data left for our analysis; which means a total of $5,050$ *disqus* and *twitter* profiles together.

## 3.2 Topic and Time Extraction

Topic is the context of plain text. Topic extractors are usually designed using different *Machine Learning* techniques. In this process, first a model is trained using a lot of data. Then, the same model is put into work to find topics lying underneath plain text. Topic Modeling is a very complex procedure and to design it without a proper level of *Machine Learning* knowledge can lead to inefficient and incorrect designs. So, we decided to use already available resources that can extract topics from plain texts.

*Python* [38] has a rich library regarding different language processing techniques named *NLTK* (natural Language Tool kit) [39]. *NLTK* has one method which identifies Nouns and Noun-phrases from plain text. In most of the cases, the topic we are looking for is a Noun or Noun-phrase. Thus, we decided to use this method to extract our topics.

Then, we discarded the words which cannot be a topic. This was done manually. After this, we designed another algorithm that extracts the related words of the topics. This algorithm basically extracted all the words in a sentence except the topic word. These words are called *Topic Related Words*. Topic related words were later used to calculate the sentiment of a user towards that topic.

Let us consider the sentence : *The Titanic was interesting but lengthy*. In this sentence, the topic will be the word *Titanic*. And the related words will be {the, was, interesting, but, lengthy}. From this list the words *was*, *the* and *but* are excluded. The remaining list is : {interesting, lengthy}. These two words have a certain level of positivity and negativity within them. So, they can be used in the sentiment extraction phase.

We also extracted the time of a particular post in a specific format and saved them in *json* files for further analysis. The saved time had six components namely year, month, day, hour, minute, second. So, we saved it in the format :

$$time = \{yyyy, mm, dd, hh, mm, ss\} \quad (3.1)$$

For example if the time of a post is : 07:19:35 PM, 31 August, 2014, we saved it like :

$$time = \{2014, 08, 31, 19, 19, 35\} \quad (3.2)$$

## 3.3 Sentiment Extraction

Sentiment is one's vibe towards a particular topic. Sentiment can be completely positive or negative, or can have a mixed proportion of positive and negative vibe, or can be completely neutral. Extracting one's emotion from plain text can be a complex task. *Python* [38] has a *Sentiment Intensity Analyzer* [40] which can give three scores to one's vibe towards a topic. Three scores are based on how positive, negative or neutral one is towards that particular topic. We extracted and saved these sentiment scores in *json* files for further analysis. *Sentiment Intensity Analyzer* receives a bag of words an gives back the output score.

For example, consider the Titanic example of above. The set {interesting, lengthy} is passed to the analyzer and it returns a polarity (positive, negative, neutral) and an polarity intensity score from $0 - 1$. This score is later used in our scoring process.

# Chapter 4

# Our Approach

## 4.1 Overview

The goal of this thesis is to identify the profile of a person in some site from a given list, when another profile of the same person in some other site is given. In this section, we will present a formal definition of the problem ahead of us. In addition, we will discuss the basic idea of our model dealing with this problem.

Let us consider two social media sites X and Y. A person has account on both these sites. In site X, the person has account, $A_x$ and in site Y, this person has account $A_y$. We know about $A_x$ and everything available in it; our task is to identify $A_y$ from a probable list. This probable list is given as *input* to our model along with $A_x$. We will use the topic, emotion, vibe of a post made by the user in site X along with the time of that post as our data, part one. We will do the same analysis in site Y and retrieve the data, part two. These data we get will be simple textual data. For the sake of easier comparisons, we will convert these to numeric data using available techniques. Closer the match between two data, higher the probability that they belong to the same person. To ease this match, our model gives a score to each member of the probable list. The one with maximum score is our *output*.

In our approach, we will use *disqus* and *twitter* as the two social media sites. We will use two different methods in our analysis. First of those will not consider the timings of the post made by an user in his *disqus* or *twitter* account. This model will be called *Non-temporal Model*. Secondly, we will start considering the timings of the posts to find how significant an effect these timings can play in the matching process. We name the later model *Temporal Model*.

For the *Non-temporal Models*, our idea is to consider the whole timespan a single bin and put all the topics inside that bin before we move into any comparison. As for the comparison, we will compare the *probabilistic distributions* of each topic for two users.

## 4.2 Topic Based Model

In case of the *Topic Only Model*, we will ignore both the times and sentiments of the topics posted by a user. Let us consider two user profiles $U_1$ and $U_2$ where $U_1$ is a *disqus* user and $U_2$ is a *twitter* user. $U_1$ and $U_2$ respectively has topics $T_1 = \{t_{11}, t_{12}, \ldots, t_{1n}\}$ and $T_2 = \{t_{21}, t_{22}, \ldots, t_{2m}\}$.

We get the combined topic set $T = T_1 \cup T_2$. And for each topic $t_i \in T$, we will find the probability of a user, $U$ discussing that topic =

$$P(t_i|U) = \frac{i}{N} \qquad (4.1)$$

where, $i$ is the number of times topic $t_i$ appeared in user $u$'s posts and

$$N = \sum_{i=1}^{T_{size}} i \qquad (4.2)$$

Now, we will get the similarity on topics between $U_1$ and $U_2$:

$$Sim_t = S_t * \log(S_t) \qquad (4.3)$$

where,

$$S_t = \sum_i [P(t_i|U_1) * P(t_i|U_2)] \qquad (4.4)$$

Here, Logarithm is introduced as the values of $S_t$ can be very low. Here, $Sim_t$ will be the final score deciding ranks.

## 4.3 Sentiment Based Model

In case of the *Sentiment Only Model*, we will ignore both the topic frequencies and times. Let us consider two user profiles $U_1$ and $U_2$ where $U_1$ is a *disqus* user and $U_2$ is a *twitter* user. $U_1$ and $U_2$ respectively has topics $T_1 = \{t_{11}, t_{12}, \ldots, t_{1n}\}$ and $T_2 = \{t_{21}, t_{22}, \ldots, t_{2m}\}$.

We declare $S$ as the set of all possible sentiments.

$$S = \{Positive, Negative, Neutral\} \qquad (4.5)$$

We get the combined topic set $T = T_1 \cup T_2$. And for each topic $t_i \in T$ and each sentiment $s_j \in S$, we will find the probability of a user, $U$ discussing that topic with that sentiment

$$= P(t_i, s_j | U)$$

Now, we will get the similarity on topics between $U_1$ and $U_2$:

$$Sim_s = S_s * \log(S_s) \tag{4.6}$$

where,
$$S_s = \sum_i \sum_j [P(s_j|U_1) * P(t_i, s_j|U_1)] * [P(s_j|U_2) * P(t_i, s_j|U_2)] \tag{4.7}$$

and, $P(s_j|U)$ is the probability that user, $U$ will post with sentiment $s_j$.

## 4.4 Combined Model

In the *Topic Only* or *Sentiment Only* models, we consider only $Sim_t$ or $Sim_s$ respectively as our final score. But, we can take their weighted average to make a combined model. How much weight each of them will get is not something that we can fix from the start. We will fix the weights through several rounds of experimenting. So, in this combined model :

$$Sim_{final} = W_1 * Sim_t + W_2 * Sim_s \tag{4.8}$$

Where, $W_1$ and $W_2$ will be calculated through experiment. and

$$W_1 + W_2 = 1 \tag{4.9}$$

## 4.5 Temporal Models

For temporal analysis, we will select a time window, $w$ and a shift variable $\tau$. Then, slice the total timespan into overlapping windows of size $w$ at $\tau$ difference. To have a better understanding, let us say $w = 15$, $\tau = 7$, total days= $210$ i.e. 7 months, then there will be 30 windows starting from day $0 - 15$, $7 - 21$, $15 - 30$ and so on.

We will calculate $Sim_t$ and $Sim_s$ for each window just like the above mentioned methods and take the average and variance of those values. But, while calculating these values for a window, only the posts that lie within our window will be considered.

However, to find the divergence between two probability distribution, our decision is to compute the *K-L Divergence*, which is defined as:

$$KL(P||Q) = \sum_i p_i * \log_2(p_i/q_i) \tag{4.10}$$

which indicates how much the distribution $P$ deviates from the distribution $Q$ taking measurement at each point, $i$. As, this is an *asymmetric* function, we will measure overall divergence by considering *bi-directional deviation* i.e.

$$KL = KL(P||Q) + KL(Q||P) \tag{4.11}$$

Just like the previous section, we can also have two different *Temporal Model*s here namely *Temporal Topic Only Model* and *Temporal Sentiment Only Model* where only topic counts in a time bin or topic sentiments in a time bin will be considered respectively.

## 4.6 Two Phase Method

This method is a temporal one. In this method, first we run analysis according to the methods described in 4.5 and 4.2. That is, we run a Topic Based Temporal analysis on the dataset. After the analysis, top 10 profiles are selected again and they are passed through a second stage just like 4.3 and 4.5. The second stage is simply a Sentiment Based Temporal Method. The output of the later stage is the final output.

Two Phase Method is more sophisticated and complex, computational time is also higher. However, the performance was not satisfactory because of the inclusion of sentiments in the later stage.

# Chapter 5

# Distance Based Scoring Model

This method was based more on intuition rather than theoretical point of view. Frequency of a topic denotes the number of time a topic has appeared in the time window in the account of the user. Normalized frequency is the proportion of this particular topic to all the topics in that window. So, normalized frequency is always less than or equal to 1. Now, any number less than 1, when squared becomes smaller. One thing that is worth mentioning is that we are not considering any time bins in this model. Which means, all the posts fall into one big time bin.

## 5.1 Intuitions

The intuitions that led to the decision of making this model are listed below:

  i. If one topic is existent in *disqus* and non existent in *twitter* or vice versa, we will take the absolute distance of their normalized frequency as we intend to penalize more here.

 ii. If one topic is existent both in *disqus* and in *twitter*, we will take the squared distance of their normalized frequency. Thus reducing the penalty as the topic is existent in both platforms.

iii. There are provisions for some bonus. If a topic appears both in *disqus* and in *twitter* more than a certain number of times, we will award this situation because, a topic repeatedly appearing in both accounts highly suggests that they belong to the same person.

 iv. In case of sentiment, if a topic is absent in one of the platforms, we will skip the distance calculation for this particular topic as making assumptions for absent topic in a platform does not make any sense.

## 5.2 The Algorithm

In Algorithm 1 we show how to calculate the desired score. Where $W_1$ is the frequency distance score weight and $W_2$ is the emotion distance score weight. The above mentioned intuitions are put into action for the algorithm.

---
**Algorithm 1** Calculate Distance Based Score $Part - 1$. Input : $\mathbf{W_1}, \mathbf{W_2}$
---
1: $allTopics = [disqusTopics \cup twitterTopics]$
2: $allDistinctTopics = \mathbf{set}(allTopics)$
3: $combinedDistance = 0$
4: $totalFreqDistance = 0$
5: $i = 0$
6: **for** $i < \mathbf{len}(allDistinctTopics)$ **do**
7:    $i = i + 1$
8:    **if** $len(twitterTopics) == 0$ **then**
9:       **Return** 0
10:    **else**
11:       $tVal = twitterTopicCounter[i]/\mathbf{len}(twitterTopics)$
12:    **end if**
13:    **if** $len(disqusTopics) == 0$ **then**
14:       **Return** 0
15:    **else**
16:       $dVal = disqusTopicCounter[i]/\mathbf{len}(disqusTopics)$
17:    **end if**
18:    **if** $tVal == 0$ **or** $dVal == 0$ **then**
19:       $totalFreqDistance+ = \mathbf{abs}(tVal - dVal)$
20:    **else**
21:       $totalFreqDistance+ = (tVal - dVal)^2$
22:    **end if**
23:    **if** $twitterTopicCounter[i] >= 5$ **and** $disqusTopicCounter[i] >= 3$ **then**
24:       $totalFreqDistance- = tVal * dVal * (tVal - dVal)^2$
25:    **end if**
26: **end for**
27: $combinedDistance = W_1 * totalFreqDistance$
28: $totalEmotionDistance = 0$
29: $div = 0$
30: $i = 0$
---

**Algorithm 2** Calculate Distance Based Score $Part-2$. Input : $\mathbf{W_1}, \mathbf{W_2}$

1: **for** $i < \mathbf{len}(allDistinctTopics)$ **do**
2:    $tEmotion = twitterSentiment.\mathbf{get}(i, \mathbf{pos} : 0.0, \mathbf{neg} : 0.0, \mathbf{neu} : 0.0)$
3:    $dEmotion = disqusSentiment.\mathbf{get}(i, \mathbf{pos} : 0.0, \mathbf{neg} : 0.0, \mathbf{neu} : 0.0)$
4:    **if** $tEmotion[\mathbf{pos}] == 0.0$ **and** $tEmotion[\mathbf{neg}] == 0$ **and** $tEmotion[\mathbf{neu}] == 0$ **then**
5:      continue
6:    **end if**
7:    **if** $dEmotion[\mathbf{pos}] == 0.0$ **and** $dEmotion[\mathbf{neg}] == 0$ **and** $dEmotion[\mathbf{neu}] == 0$ **then**
8:      continue
9:    **end if**
10:   $temp = 0$
11:   $temp+ = \mathbf{abs}(tEmotion[\mathbf{pos}] - dEmotion[\mathbf{pos}])$
12:   $temp+ = \mathbf{abs}(tEmotion[\mathbf{neg}] - dEmotion[\mathbf{neg}])$
13:   $temp+ = \mathbf{abs}(tEmotion[\mathbf{neu}] - dEmotion[\mathbf{neu}])$
14:   $temp/ = 3$
15:   $totalEmotionDistance+ = temp$
16:   $div+ = 1$
17:   **if** $div == 0$ **then**
18:      $combinedDistance+ = 0$
19:   **else**
20:      $combinedDistance+ = W_2 * (totalEmotionDistance/div)$
21:   **end if**
22: **end for**
23: **Return** $-combinedDistance$

## 5.3 Explanation of the Algorithm

The line by line explanation of the algorithm is presented in the table 5.1. Important lines of the algorithm are explained in bold characters. In the last line of Part-2 of the algorithm, we return the negative of the combined distance because, we want the output to be higher for closer match. We could have returned the positive combined distance. In that case, we would have to consider the reverse order to find closest match.

**Pos, Neg, Neu** stands for Positive, Negative, and Neutral respectively. **Abs** stands for absolute. *dVal, dEmotion* stands for Disqus and *tVal, tEmotion* stands for Twitter.

Table 5.1: Algorithm Explained

| Line no. | Explanation of the Code |
| --- | --- |
| **1 (Part-1)** | All the existing topics in disqus and twitter are stored into allTopics |
| 2 | Using set operation we remove the repetitions from the allTopics list |
| 3 | Initialize Combined Distance to 0 |
| 4 | Initialize Total Frequency Distance to 0 |
| 5 | Initialize looping variable $i$ |
| 6 | Loop Until $i$ reaches the number of distinct topics in disqus and twitter combined |
| 7 | Increment looping variable |
| 8 and 9 | If there is no topics in twitter, score is $0$ |
| 10 and 11 | $tVal$ is the normalized frequency of $i$th topic in twitter |
| 13 and 14 | If there is no topics in disqus, score is $0$ |
| 15 and 16 | $dVal$ is the normalized frequency of $i$th topic in disqus |
| 18 and 19 | **If a topic is absent in either twitter or disqus, take the absolute distance** |
| 20 and 21 | **If a topic is existent in both platforms, take the squared distance** |
| 23 and 24 | **If a topic appears in twitter more than $5$ times and in disqus more than $3$ times, give some bonus** |
| 27 | Multiply Total Frequency Distance by its weight, $W_1$ and add it to Total Score |
| 28, 29, 30 | Initialize different variables |
| **1 (Part-2)** | Loop Until $i$ reaches the number of distinct topics in disqus and twitter combined |
| 2, 4, 5 | If a topic is absent in twitter, sentiment calculation for this topic is skipped |
| 3, 7, 8 | If a topic is absent in disqus, sentiment calculation for this topic is skipped |
| 10-16 | Total Emotion Distance Calculation like before, all absolute value taken |
| 17 and 18 | If No topic is present in disqus or twitter, nothing is added to the distance |
| 19 and 20 | Multiply Total Emotion Distance by its weight, $W_2$ and add it to Total Score |
| 23 | Return the **negative** of the combined distance |

# Chapter 6

# Experimental Results

This chapter contains all the performance metrics, testing methodologies, and testing results.

## 6.1 Performance Metrics

There will be several metrics for performance measure. They are discussed below:

### 6.1.1 Average Rank

Let us consider, there are $N$ Disqus profiles and $N$ Twitter profiles. Also consider a Disqus profile, $P_d$. Within the $N$ candidate Twitter profiles, one belongs to the person with Disqus profile $P_d$. Let us name this Twitter profile $P_t$. Now, in the scoring process, all the $N$ Twitter profiles will be matched against $P_d$ and they will be given a score. Later, these $N$ Twitter profiles will be ranked form 1 to $N$ with ranked 1 profile having the highest match score. Now, Our target profile $P_t$ will be ranked in some place between 1 to $N$. Let us call the the rank of $P_t$ $R$.

We will have $(N-1)$ more Disqus profiles other than $P_d$ to match against the available $N$ Twitter profiles. Each time, we will get a rank $R$, which denotes the rank of the original profile after the scoring process.

Average rank means the arithmetic average of these $N$ $R$ values. So, average rank is :

$$AverageRank = \frac{\sum_{i=1}^{N} R_i}{N} \qquad (6.1)$$

**Example**

If we have $5$ Disuqs profiles in total and the ranks they get after scoring are respectively $\{3, 4, 1, 4, 2\}$, then the average rank is :

$$AverageRank = \frac{(3 + 4 + 1 + 4 + 2)}{5} \tag{6.2}$$

$$= 2.8 \tag{6.3}$$

So, from the discussion above, quite clearly we can say that, *lower the average rank, the better.*

### 6.1.2 Accuracy

Keeping the discussions of 6.1.1 in mind, we will define a new performance measure metric named *Accuracy*. When our target Twitter profile $P_t$ is ranked $1^{st}$ among the $N$ candidates, we call it a success. Accuracy the the ratio of number of success to the total number of profiles tested, which is $N$. If number of success is defined by $S$, then,

$$Accuracy = \frac{S}{N} \tag{6.4}$$

**Example**

If there are $2500$ total profiles tested and in $550$ cases, our target profile is ranked $1^{st}$, then accuracy is:

$$Accuracy = \frac{550}{2500} \tag{6.5}$$

$$= 0.22 \tag{6.6}$$

$$= 22\% \tag{6.7}$$

We can say that, *Higher the accuracy, the better*

### 6.1.3 Top-K Positions

Taking the discussion of 6.1.1 into consideration, we define Top-K Positions as the number of candidate profiles ranked less than or equal to $K$. Here $K$ is an integer.

**Example**

If $1150$ profiles out of total $2500$ are ranked within $\{1, 2, 3\}$, then the Top-3 is:

$$Top_3 = \frac{1150}{2500} \tag{6.8}$$

$$= 0.46 \tag{6.9}$$

$$= 46\% \tag{6.10}$$

if $1550$ profiles are ranked within $\{1, 2, 3, 4, 5\}$, then the Top-5 is:

$$Top_5 = \frac{1550}{2500} \tag{6.11}$$

$$= 0.62 \tag{6.12}$$

$$= 62\% \tag{6.13}$$

if $1850$ profiles are ranked within $\{1, 2, 3, 4, 5, 6, 7, 8, 9, 10\}$, then the Top-10 is:

$$Top_{10} = \frac{1850}{2500} \tag{6.14}$$

$$= 0.74 \tag{6.15}$$

$$= 74\% \tag{6.16}$$

## 6.2 Testing Methodologies and Results

$50$ Disqus and $50$ Twitter profiles are tested here for each of the method.

## 6.2.1 Topic Based Model (Non-Temporal)

In this model, the whole time-space was considered just one big bin. We just used the method described in 4.2 to calculate the score.

Table 6.1: Results of Topic Based Model (Non-temporal)

| Average Rank | First (Rank=1) | Top-3 | Top-10 |
|---|---|---|---|
| 3.4 | 26 | 34 | 45 |

## 6.2.2 Sentiment Based Model

This model by itself, described in 4.3 is massively under-performing. Thus, we excluded further testing for this model.

## 6.2.3 Topic Based Model (Temporal)

In this model, the whole time-space was divided into some overlapping bins. This method was carried out as per the technique described in 4.5.

Table 6.2: Results of Topic Based Model (Temporal)

| Window Size (w) | Shifting Amount ($\tau$) | Average Rank | First (Rank=1) | Top-3 | Top-10 |
|---|---|---|---|---|---|
| 365 | 365 | 3.42 | 25 | 32 | 46 |
| 365 | 180 | 3.52 | 23 | 32 | 46 |
| 365 | 90 | 3.54 | 22 | 32 | 46 |
| 180 | 90 | 3.56 | 22 | 32 | 46 |
| 730 | 365 | 3.42 | 25 | 32 | 46 |
| 730 | 180 | 3.54 | 23 | 33 | 46 |
| 1460 | 730 | 3.06 | 28 | 35 | 47 |

The results of Table 6.2 are illustrated in Figure 6.1. The blue bars indicate the Window Size. The red bars indicate the Shifting Amount. The green line represents the Average Rank with respective Window Size and Shifting Amount. The lowest Average Rank is obtained for $w = 1460$ and $\tau = 730$. For the other scenarios, the Average Rank changes quite irregularly and does not seem to depend much on Shifting Amount. However, the higher the Window Size, the better is the Average Rank, as can be seen from Figure 6.1.

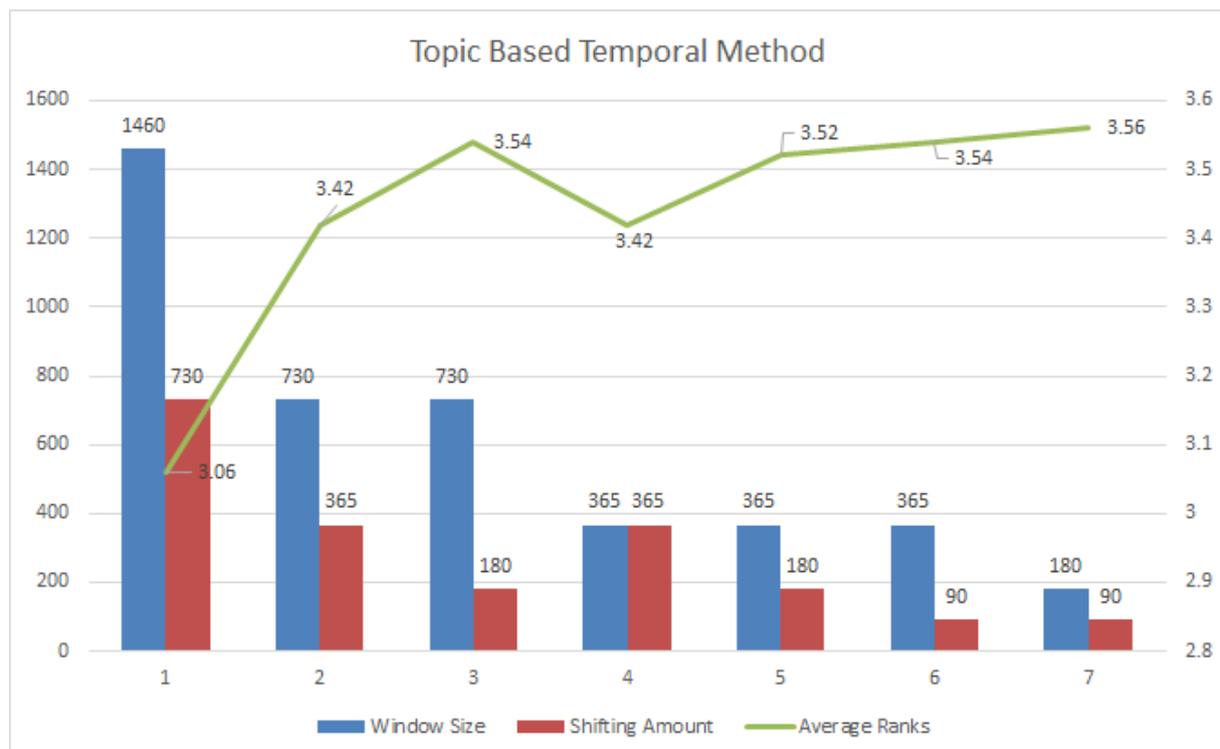

Figure 6.1: Effect of Window Size and Shifting Amount on Topic Based Model

### 6.2.4 Two Phase Method (Temporal)

The testing was performed according to 4.6.

Table 6.3: Results of Two Phase Method

| Window Size (w) | Shifting Amount ($\tau$) | Average Rank | First (Rank=1) | Top-3 | Top-10 |
|---|---|---|---|---|---|
| 365 | 365 | 4.54 | 15 | 28 | 46 |
| 730 | 365 | 4.54 | 15 | 28 | 46 |
| 365 | 180 | 4.44 | 15 | 27 | 46 |
| 730 | 180 | 4.34 | 16 | 29 | 46 |
| 365 | 90 | 4.47 | 15 | 28 | 46 |

The results of Table 6.3 are illustrated in Figure 6.2. The blue bars indicate the Window Size. The red bars indicate the Shifting Amount. The green line represents the Average Rank with respective Window Size and Shifting Amount. The lowest Average Rank is obtained for $w = 730$ and $\tau = 180$. For the other scenarios, the Average Rank changes quite irregularly and does not seem to depend much on the Average Rank or the Shifting Amount, as can be seen from Figure 6.1.

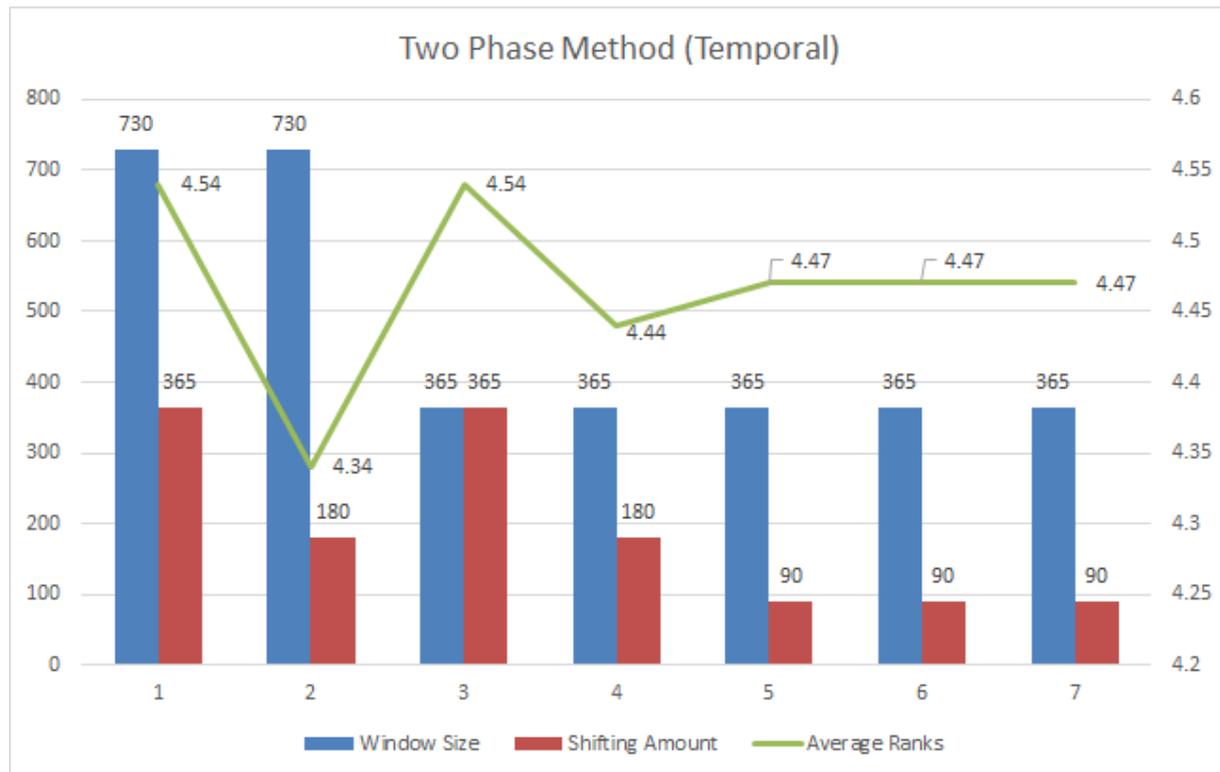

Figure 6.2: Effect of Window Size and Shifting Amount on Two Phase Method

## 6.3 Comparison of the Results

### 6.3.1 Topic Based Temporal Method Vs. Two Phase Method

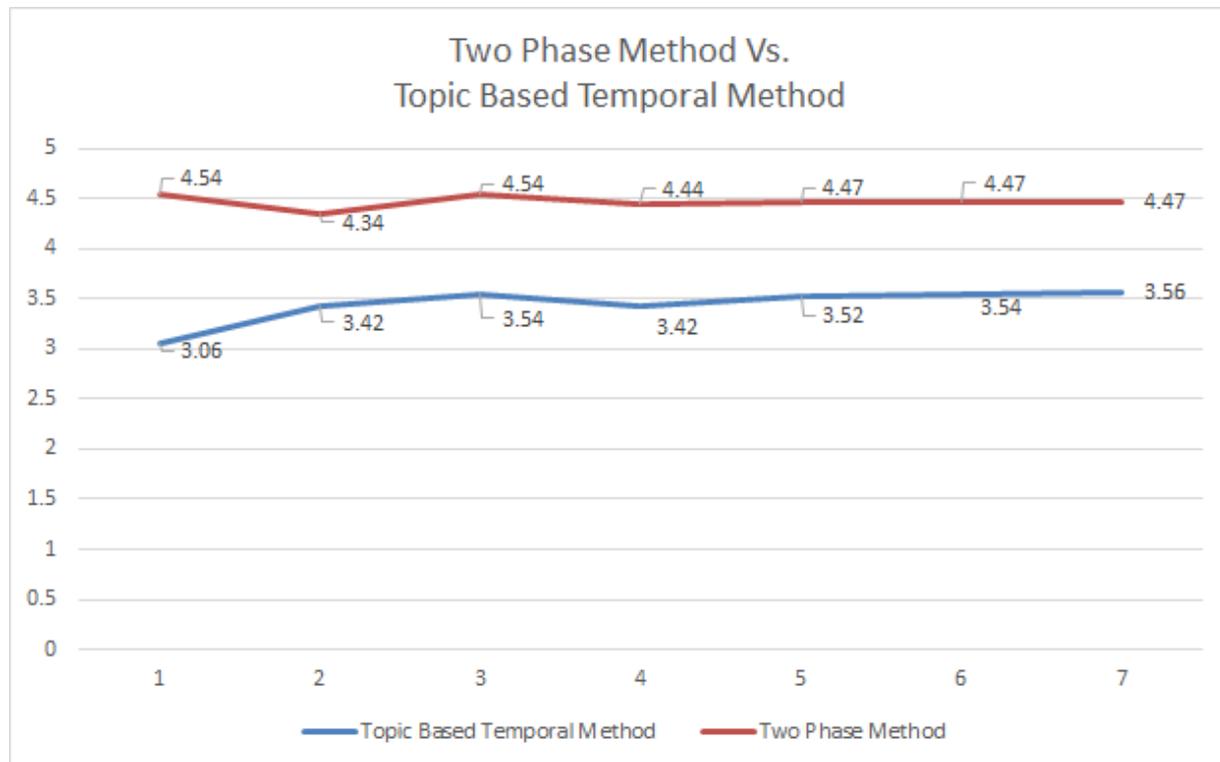

Figure 6.3: Average Rank Comparison of Topic Based Temporal Method and Two Phase Method

An Average Rank comparison of Topic Based Temporal Method and Two Phase Method is shown in Figure 6.3. The blue line indicates the Topic Based Temporal Method and the red line indicates the Two Phase Method. As we can see from Figure 6.3, the Topic Based Temporal Method is a clear winner. Two Phase Method incorporates sentiment analysis in its second step. This is why the performance of Two Phase Method was worse.

### 6.3.2 Topic Based Temporal Method Vs. Non-temporal Method

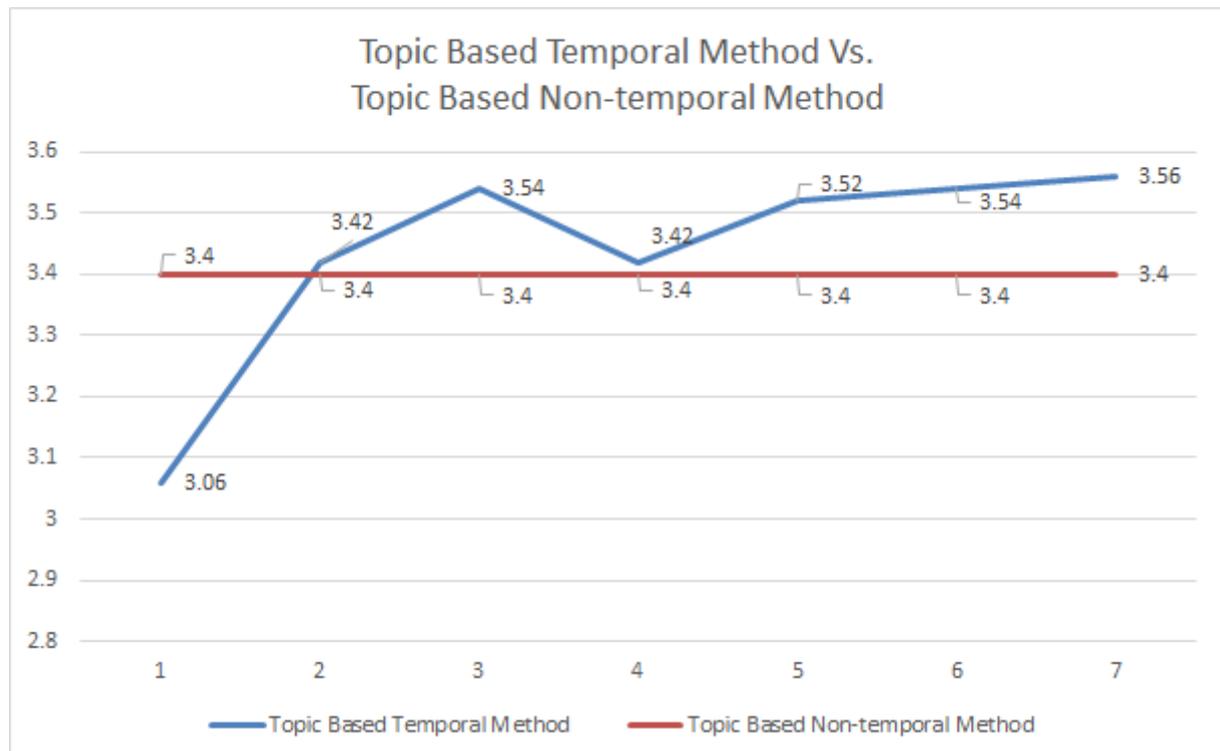

Figure 6.4: Average Rank Comparison of Topic Based Temporal Method and Non-temporal Method

An Average Rank comparison of Topic Based Temporal Method and Topic Based non-temporal Method is shown in Figure 6.4. The blue line indicates the Topic Based Temporal Method and the red line indicates the Topic Based Non-Temporal Method. As we can see from Figure 6.4, we cannot really pick a clear winner. The temporal method performed much better at one stage and almost followed the non-temporal method in the other stages. So, roughly speaking, the temporal method is slightly better than the Non-temporal method.

### 6.3.3 Topic Based Non-temporal Method Vs. Two Phase Method

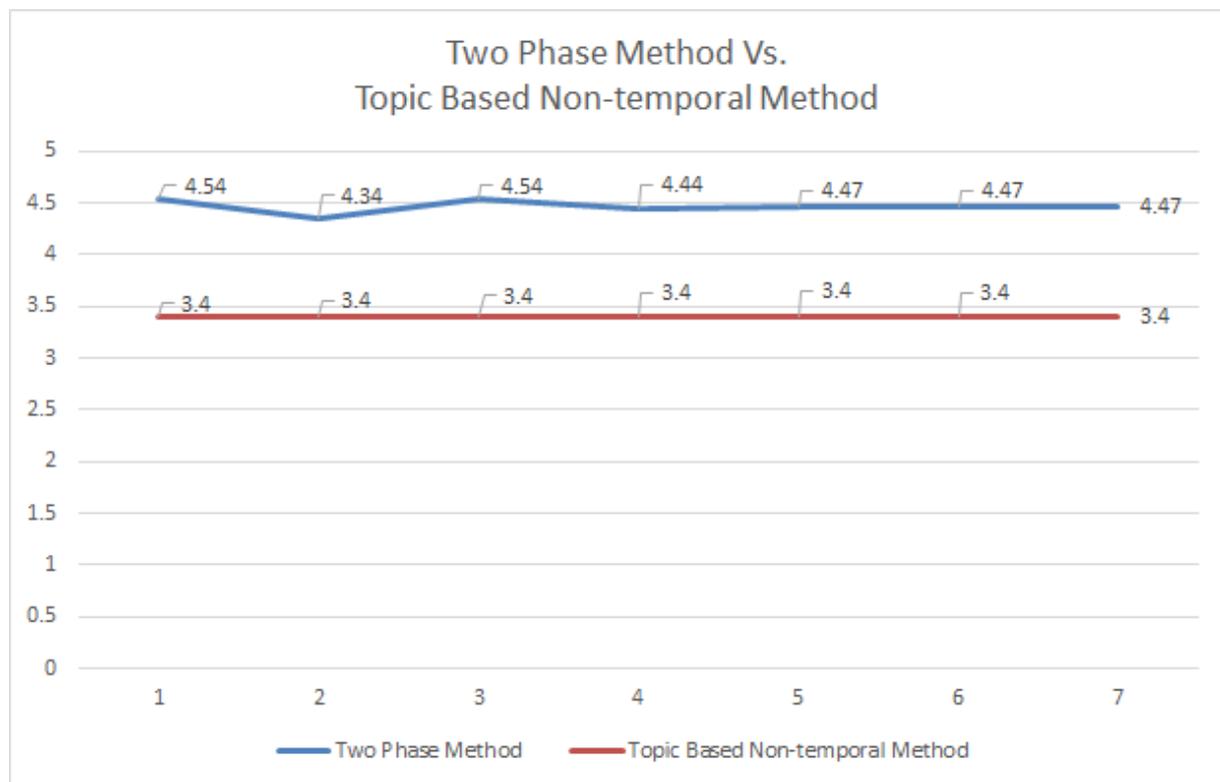

Figure 6.5: Average Rank Comparison of Topic Based Non-temporal Method and Two Phase Method

An Average Rank comparison of Topic Based Non-temporal Method and Two Phase Method is shown in Figure 6.5. The red line indicates the Topic Based Non-temporal Method and the blue line indicates the Two Phase Method. As we can see from Figure 6.5, the Topic Based Non-temporal Method is a clear winner. Two Phase Method incorporates sentiment analysis in its second step. This is why the performance of Two Phase Method was worse.

## 6.4 Testing Results of Distance Based Scoring Model

### 6.4.1 Methodologies

Here, we tested 2525 Disqus profiles and 2525 Twitter profiles. The algorithm 1 was used and it was significantly faster than the procedures in the above methods.

### 6.4.2 Results

We considered the weight of topic frequency, $W_1 = 0.75$ and weight of sentiment $W_2 = 0.5$.

The results are listed below:

| Accuracy | 24.198% |
|---|---|
| Top-3 | 34.2574% |
| Top-10 | 47.802% |
| Average Rank | 158.217 |

An Accuracy of $24.198\%$ and an Average Rank of $158.217$ out of $2525$ makes this model a well performing one.

## 6.5 Analysis and Limitations the Thesis

The models we have built are not free of flaws. The flaws and their probable causes along with a complete analysis are given below.

i. **Sentiment Analysis is Less Useful:** As we saw in the discussion and analysis in this chapter, whenever we incorporated sentiment analysis the results got worse. There can be only one reason behind this, which is the sentiments we extracted were not entirely correct. These sentiments were calculated based on *Topic related Words*. Thus, the sentiment for all the topics in a sentence were the same. This is a flaw. This is the reason, the sentiments do not represent the actual attributes of the profile and do not bring any good output.

ii. **Temporal Analysis is not Strictly Better:** As can be seen from the analysis in this chapter, temporal analysis is not always better. But, when it lags behind non-temporal method, the margin is extremely low. On the other hand, when it wins, the margin of winning remains quite large. Thus, we can say that, temporal analysis is promising and useful but does not always guarantee better results.

iii. **Results are Independent of Shifting Amount:** Analysis of the results showed it clearly that the accuracy or the average rank is not dependent on the Shifting Amount.

iv. **Results are Dependent on Window Size:** The results, although independent of the Shifting Amount, were dependent on the Window Size. The larger the window size we took, the better the results got.

# Chapter 7

# Conclusion and Future Direction

In this thesis, our task was to build an identity resolver model based on the topic, emotion and timings of the posts made by users in social media. We have made several models. There have been models where we have not considered the timings and we have called them non-temporal models. There have been models, where we have considered the timings and have called them temporal models. In a comparison of temporal and non-temporal models, neither has been found strictly better but we have concluded that the temporal models have been roughly better. Sentiment based analysis has also been performed but they have not been impressing. So, we have excluded sentiment analysis for the later sections of our experiment. The reason we have found behind the under-performing sentiment analysis has been the process how they have been extracted. In case of temporal models, the results have been independent of the Shifting Amount $\tau$ but have depended more or less on the Window Size $w$. We have also experimented with a distance based reward and punishment type scoring model; which by itself, has performed quite well with and accuracy of $24.198\%$ and an average rank of $158.217$ out of $2525$ which has been very impressive.

There can be several direct extensions to the work we have done. Firstly, the sentiment analysis can be made better. The problem we had with sentiment analysis in this thesis was the extraction procedure. We marked the same sentiment on every topic available in a sentence which was the flaw. We will change this extraction procedure by individually calculating sentiments for each topic and not as a sentence. Secondly, the temporal analysis will be extended by adding another one or two phases. As we saw in the results of this thesis, many of the profiles stay within top-10, although they do not come first. An extra round or two of analysis of these $10$ profiles will increase the Accuracy to some extent. Finally, we will try to improve the distance based scoring model by readjusting the weights and changing the reward of punishment given to see what makes the best possible results.